%% file: main.tex
\documentclass[sigconf]{acmart}
\AtBeginDocument{%
  }

\setcopyright{acmlicensed}
\copyrightyear{2026}
\acmYear{2026}
\acmDOI{10.1145/3774904.3792572}
\acmConference[WWW '26]{the ACM WEB Conference}{April 13--17,
  2026}{Dubai, United Arab Emirates}
\acmISBN{978-1-4503-XXXX-X/2018/06}

\usepackage{multirow}
\usepackage{algorithm}
\usepackage{algorithmic}
\begin{document}

\title{E2PL: Effective and Efficient Prompt Learning for Incomplete Multi-view
Multi-Label Class Incremental Learning}


\author{Jiajun Chen}
\affiliation{%
  \institution{Zhejiang University}
  \department{Center for Data Science}
  \city{Hangzhou}
  \state{Zhejiang}
  \country{China}
}
\email{chenjjcccc@zju.edu.cn}

\author{Yue Wu}
\affiliation{%
  \institution{Ant Group}
  \city{Hangzhou}
  \state{Zhejiang}
  \country{China}
}
\email{yuyue.wy@antgroup.com}

\author{Kai Huang}
\affiliation{%
  \institution{Ant Group}
  \city{Hangzhou}
  \state{Zhejiang}
  \country{China}
}
\email{kevin.hk@antgroup.com}

\author{Wenxi Zhao}
\affiliation{%
  \institution{Zhejiang University}
  \department{School of Software Technology}
  \city{Ningbo}
  \state{Zhejiang}
  \country{China}
}
\email{2840126Z@student.gla.ac.uk}

\author{Yangyang Wu}
\authornote{Corresponding author.}
\affiliation{%
  \institution{Zhejiang University}
  \department{School of Software Technology}
  \city{Ningbo}
  \state{Zhejiang}
  \country{China}}
\email{zjuwuyy@zju.edu.cn}

\author{Xiaoye Miao}
\affiliation{%
 \institution{Zhejiang University}
  \department{Center for Data Science}
 \city{Hangzhou}
  \state{Zhejiang}
 \country{China}}
\email{miaoxy@zju.edu.cn}

\author{Mengying Zhu}
\affiliation{%
  \institution{Zhejiang University}
  \department{School of Software Technology}
  \city{Ningbo}
  \state{Zhejiang}
  \country{China}}
\email{mengyingzhu@zju.edu.cn}

\author{Meng Xi}
\affiliation{%
  \institution{Zhejiang University}
  \department{School of Software Technology}
  \city{Ningbo}
  \state{Zhejiang}
  \country{China}}
\email{ximeng@zju.edu.cn}

\author{Guanjie Cheng}
\affiliation{%
  \institution{Zhejiang University}
  \department{School of Software Technology}
  \city{Ningbo}
  \state{Zhejiang}
  \country{China}}
\email{zhaoxinkui@zju.edu.cn}

\renewcommand{\shortauthors}{Trovato et al.}

\begin{abstract}
Multi-view multi-label classification (MvMLC) is indispensable for modern web applications aggregating information from diverse sources. However, real-world web-scale settings are rife with missing views and continuously emerging classes, which pose significant obstacles to robust learning.
Prevailing methods are ill-equipped for this reality, as they either lack adaptability to new classes or incur exponential parameter growth when handling all possible missing-view patterns, severely limiting their scalability in web environments.
To systematically address this gap, we formally introduce a novel task, termed \emph{incomplete multi-view multi-label class incremental learning} (IMvMLCIL), which requires models to simultaneously address heterogeneous missing views and dynamic class expansion.
To tackle this task, we propose \textsf{E2PL}, an Effective and Efficient Prompt Learning framework for IMvMLCIL.
\textsf{E2PL} unifies two novel prompt designs: \emph{task-tailored prompts} for class-incremental adaptation and \emph{missing-aware prompts} for the flexible integration of arbitrary view-missing scenarios.
To fundamentally address the exponential parameter explosion inherent in missing-aware prompts, we devise an \emph{efficient prototype tensorization} module, which leverages atomic tensor decomposition to elegantly reduce the prompt parameter complexity from exponential to linear w.r.t. the number of views.
We further incorporate a \emph{dynamic contrastive learning} strategy explicitly model the complex dependencies among diverse missing-view patterns, thus enhancing the model's robustness.
Extensive experiments on three benchmarks demonstrate that \textsf{E2PL} consistently outperforms state-of-the-art methods in both effectiveness and efficiency.
The codes and datasets are available at https://anonymous.4open.science/r/code-for-E2PL.

\end{abstract}

\ccsdesc[500]{Computing methodologies~Supervised learning by classification}
\keywords{Incomplete multi-view data, multi-label classification, class incremental learning, prompt learning, contrastive learning}


\maketitle

\input{1.Introduction}

\input{2.Related-Work}

\input{3.Method}

\input{4.Experiment}

\input{5.Conclusion}

\balance
\bibliographystyle{ACM-Reference-Format}
\bibliography{main}

\appendix
\input{6.Appendix}

\end{document}

%% file: 1.Introduction.tex
\begin{figure}[t]
\centering
  \includegraphics[width=\columnwidth]{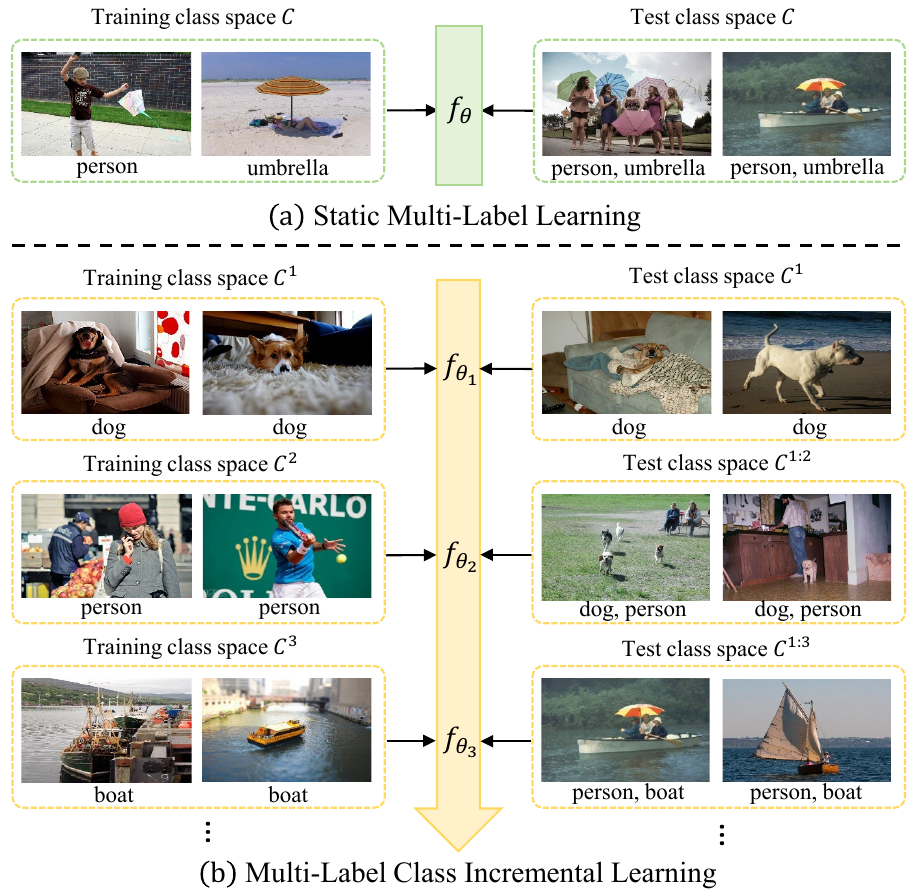}
  \caption{(a) Illustration of Static Multi-Label Learning, where the model is trained and evaluated on a fixed set of classes. (b) Illustration of Multi-Label Class Incremental Learning, where new classes (e.g., “dog”, “person”, “boat”) are introduced sequentially, requiring the model to adapt to an expanding class set while retaining prior knowledge.}
  \label{fig:static}
  \vspace{-0.2in}
\end{figure}

\label{sec:Introduction}
Multi-view multi-label classification (MvMLC) has emerged as a cornerstone technique for web applications that seek to integrate heterogeneous information from multiple sources and capture intricate semantic dependencies among labels~\cite{zhang2018latent, zhao2022non, li2025semi}.
By combining distinct feature representations across views, MvMLC not only enhances predictive accuracy but also increases robustness compared to its single-view counterparts, enabling a wide range of applications such as multimedia retrieval, medical diagnosis, and remote sensing.
In practice, however, multi-view data collected from real-world web environments are often incomplete due to sensor malfunctions, occlusions, user privacy restrictions, or prohibitive acquisition costs. 
This leads to the incomplete multi-view multi-label classification (IMvMLC) problem, which has attracted significant attention for its practical importance and inherent complexity~\cite{liu2023dicnet,liu2025reliable, wen2023deep}.
Despite notable progress, existing IMvMLC solutions still encounter two core obstacles that hinder large-scale deployment in dynamic web systems:

\emph{Class-Incremental Flexibility Barrier (\textbf{CH1}):} The prevailing paradigm~\cite{zhao2022learning, liu2023label, liu2024attention} is grounded in static multi-label learning, where the mapping from input features to labels is constructed statically and assumed fixed during deployment as depicted in Figure~\ref{fig:static} (a). However, in open-world web applications, the continual introduction of new categories is the norm rather than the exception. 
For instance, in medical image analysis, the discovery of novel disease subtypes is an ongoing process due to advances in research and the accumulation of clinical data \cite{kumar2018example, chen2024classifier}. This necessitates that deployed models not only retain knowledge of previously seen categories, but also incrementally incorporate and recognize new classes as they arise, often without access to the original training data and while avoiding catastrophic forgetting illustrated in Figure~\ref{fig:static} (b). Unfortunately, existing IMvMLC methods, which are typically built on static learning assumptions, lack the flexibility to efficiently accommodate such incremental categories. This incremental inadaptability poses a critical barrier, severely limiting the practical deployment of these models in evolving web environments.

\emph{Scalability and Parameter Explosion (\textbf{CH2}):} Existing IMvMLC methods predominantly rely on complex network architectures, such as autoencoder-based frameworks \cite{liu2023masked, li2025semi, liu2024partial}, which, while effective in handling missing data, often incur high computational overhead and challenging optimization problems. Their rigid designs fundamentally limit scalability and adaptability in scenarios with continuously expanding categories. Recently, prompt-based methods \cite{lee2023multimodal, zheng2024decomposed, lang2025retrieval} have been proposed as lightweight and potentially scalable alternatives for incomplete data learning. However, these methods require designing specific prompt templates for all possible missing-view patterns, resulting in exponential complexity, specifically $\mathcal{O}(2^n)$, where $n$ is the number of views and $2^n$ corresponds to all possible missing-view patterns. This inevitably leads to a parameter explosion.

To address these critical challenges, we introduce a novel task of \emph{incomplete multi-view multi-label class incremental learning} (IMvMLCIL), which demands models to jointly manage heterogeneous missing views and dynamically expanding label spaces. To tackle this, we present \textsf{E2PL}—an Effective and Efficient Prompt Learning framework that unifies both \emph{task-tailored prompts} (TTPs) for swift adaptation to new classes and \emph{missing-aware prompts} (MAPs) for flexible integration of arbitrary missing-view patterns. 
To address the associated parameter explosion in MAPs, we introduce an \emph{efficient prototype tensorization} (EPT) module based on atomic tensor decomposition, compressing the prompt space from exponential to linear complexity. Furthermore, the \emph{dynamic contrastive learning} (DCL) strategy is employed to capture nuanced interrelations among diverse missing-view patterns, further enhancing robustness and adaptability.
In summary, our paper makes several notable contributions:

\begin{itemize}
    \item Towards the IMvMLCIL task, we propose a prompt-based framework \textsf{E2PL}, effectively addressing the challenges of heterogeneous missing views and dynamically expanding label spaces.
    To the best of our knowledge, this is the first work to holistically address the IMvMLCIL problem within a unified framework.
    \item We design an \emph{efficient prototype tensorization} module via atomic tensor decomposition, reducing the parameter complexity of missing-aware prompts from exponential $\mathcal{O}(2^n)$ to linear $\mathcal{O}(n)$ w.r.t.\ the number of views.
    \item We propose a \emph{dynamic contrastive learning} strategy to capture nuanced interrelations among diverse missing-view patterns, further enhancing the model’s robustness and adaptability in incremental learning scenarios.
    \item Extensive experiments on three benchmark datasets, demonstrating that \textsf{E2PL} consistently outperforms state-of-the-art methods in both effectiveness and efficiency.
\end{itemize}

%% file: 2.Related-Work.tex
\section{Related Work}
\label{sec:Related}

\subsection{Multi-View Multi-Label Classification}

Multi-view multi-label classification (MvMLC) has emerged as a pivotal research direction, situated at the intersection of multi-view learning and multi-label classification.
Its recent prominence is largely propelled by significant advances in multi-view representation learning.
Zhang et al.~\cite{zhang2018latent} introduced a matrix factorization approach to align latent representations across views.
\citet{wu2019multi} proposed SIMM to jointly exploit a shared subspace and captureview-specific features through adversarial, multi-label lossand orthogonal constraints. 
Zhao et al.~\cite{zhao2021consistency} presented CDMM, which learns independent predictions for each view and maximizes feature-label dependence via the Hilbert–Schmidt Independence Criterion. The challenge of data incompleteness in both features and labels adds further complexity.
Incomplete Multi-view Multi-label Classification (IMvMLC), first addressed by Tan et al.~\cite{tan2018incomplete}, projects incomplete data into a common subspace and bridges it to the label space with a projection matrix and learnable label correlations. Subsequent work has improved robustness and representation: DICNet~\cite{liu2023dicnet} employs autoencoders and instance-level contrastive learning; TSIEN~\cite{tan2024two} disentangles task-relevant and irrelevant information using information bottleneck theory; and DRLS~\cite{yan2025incomplete} separates and exploits consistent and view-specific representations via disentangled learning. Nevertheless, \emph{class-incremental flexibility barrier (CH1)} and \emph{scalability and parameter explosion (CH2)} remain insufficiently addressed, hindering the real-world deployment in dynamic environments.

\begin{figure*}[h]
  \includegraphics[width=\textwidth]
  {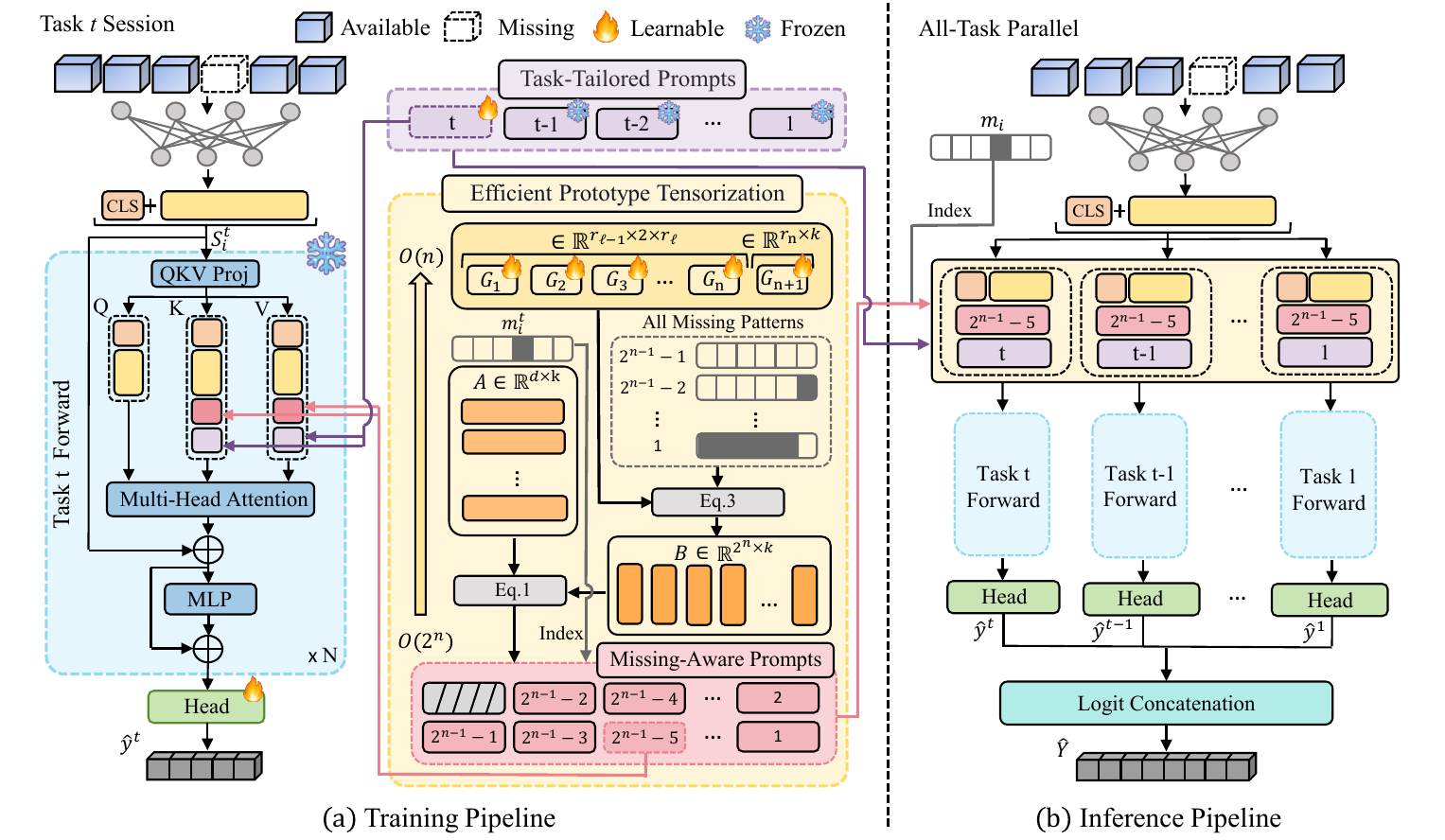}
  \caption{The overall architecture of our method. (a) Training pipeline for Task~$t$: Incomplete multi-view data undergoes linear projection and [CLS] token concatenation. Task-Tailored Prompts and Missing-Aware Prompts are added to the $\mathrm{K}$ and $\mathrm{V}$ vectors, followed by multi-label classification via the classification head. Missing-Aware Prompts are generated using the Efficient Prototype Tensorization module. (b) Inference pipeline: Forward computation for all tasks is performed in parallel, and final predictions are obtained by logit concatenation.}
  \label{fig:architecture}
  \vspace*{-0.15in}
\end{figure*}

\subsection{Prompt Learning}

Prompt learning~\cite{lester2021power, liang2022modular, khattak2023maple} adapts pre-trained models to diverse downstream tasks by designing effective prompts. Recently, the integration of prompt learning techniques into multimodal learning under the general missing modality scenario---where each specific missing pattern is regarded as a distinct task---has emerged as a lightweight and efficient adaptation strategy \cite{lee2023multimodal, guo2024multimodal, lang2025retrieval}. Despite significant progress, existing approaches typically require separate prompts for each missing modality pattern, leading to exponential growth in parameters as the number of modalities increases. \citet{jang2024towards} mitigate this by introducing Modality-Specific Prompts, reducing parameter complexity from exponential to linear by assigning independent prompts to each modality. Building on this, \citet{chen2024epe} proposed EPE-P, which further improves efficiency and performance by integrating cross-modal prompts. 
However, as the number of modalities increases, such independent modeling approaches fail to capture the nuanced dependencies across missing scenarios, ultimately limiting their adaptability and expressiveness.
In this work, we retain the independent prompt design for each missing case and further propose a parameter decomposition technique that minimizes parameter overhead while achieving optimal performance.

%% file: 3.Method.tex
\section{Methodology}
\label{sec:Overview}

\subsection{Problem Definition}

Assume there are a total of $T$ incremental sessions $\{ (\mathcal{X}^1, \mathcal{Y}^1, \mathcal{C}^1), \cdots \\ (\mathcal{X}^T, \mathcal{Y}^T, \mathcal{C}^T) \}$, where each session $t$ provides a training set $(\mathcal{X}^t, \mathcal{Y}^t)$ and a newly introduced, disjoint label set $\mathcal{C}^t$ (i.e., $\mathcal{C}^m \cap \mathcal{C}^n = \emptyset$ for $m \neq n$). For each session, $\mathcal{X}^t = \left\{ \left( x_{i,1}^t, \cdots, x_{i,n}^t \right), m_i^t \right\}_{i=1}^{N_t}$ denotes the multi-view features of $N_t$ samples, where $n$ is the number of views, $x_{i,v}^t$ is the feature vector of the $v$-th view, and $m_i^t \in \{0,1\}^n$ is a view-indicator vector with $m_{i,v}^t=1$ if the $v$-th view is available and $0$ otherwise, allowing for missing views. The corresponding label set $\mathcal{Y}^t = \{ y_i^t \}_{i=1}^{N_t}$ contains multi-label annotations for the current session, where each $y_i^t \in \{0,1\}^{|\mathcal{C}^t|}$.
The goal of \emph{incomplete multi-view multi-label class incremental learning} (IMvMLCIL) is to incrementally train a unified multi-label classification model that can accurately predict labels over all encountered classes for samples with missing multi-view features.
At each session $t$, only the current training data $(\mathcal{X}^t, \mathcal{Y}^t)$ are available for model update. After session $t$, the model is evaluated on a test set covering all classes seen so far, i.e., the cumulative class space $\mathcal{C}^{1:t} = \bigcup_{\tau=1}^{t} \mathcal{C}^\tau$.
For each test sample, the model must predict a multi-label vector over $\mathcal{C}^{1:t}$ based on its possibly incomplete multi-view features. During preprocessing, missing values in $\mathcal{X}$ are imputed with zeros.

\subsection{Prompt Design}

Existing IMvMLC methods \cite{liu2023masked, li2025semi, liu2024partial} mainly use autoencoder-based models to extract unimodal and multimodal features, which are fused and classified by multi-layer perceptrons.
While effective, these architectures are often cumbersome and lack scalability and adaptability, especially in class-incremental scenarios common to real-world applications. To address these limitations, we propose a unified prompt-based framework that leverages the efficiency and flexibility of prompt learning, as illustrated in Figure~\ref{fig:architecture}.
Our proposed framework \textsf{E2PL} orchestrates two complementary prompt types—task-tailored for class increments and missing-aware for data incompleteness—to achieve robust performance in dynamic and imperfect data environments.

\emph{Task-Tailored Prompts} (TTPs).
To support class incremental learning scenario, we structure the problem as a sequence of tasks, each bringing in new classes. 
For each task, we design TTPs specific to its requirements.
Formally, let $P_T = \{P_T^1, P_T^2, \cdots, P_T^T\}$, where $P_T^t$ denotes the prompt uniquely learned for the $t$-th task. These prompts are attached to a frozen, pre-trained transformer backbone, allowing efficient injection of task-specific knowledge without retraining the model. During training for the $t$-th task, only $P_T^t$ is updated, while prompts from previous tasks remain fixed to prevent catastrophic forgetting illustrated in Figure~\ref{fig:architecture} (a).
At inference (Figure~\ref{fig:architecture} (b)), all task-tailored prompts are processed in parallel, and the outputs from the corresponding task-specific classification heads are concatenated for final prediction. 

\emph{Missing-Aware Prompts} (MAPs).
To address the challenge posed by varying missing-view configurations in incomplete multi-view data, we propose Missing-Aware Prompts (MAPs), where each prompt is explicitly engineered to represent a unique pattern of view availability. For $n$ input views, there exist $2^n-1$ possible missing-view patterns, as each view can be either present or absent (for convenience, we use $2^n$ in the following discussion). Ideally, each pattern would be assigned a unique prompt, denoted as $P_M = \{P_M^1, \cdots, P_M^{2^n}\}$, where $P_M^i$ corresponds to the $i$-th missing-view pattern.
\vspace*{-0.1in}
\subsection{Efficient Prototype Tensorization}

The direct parameterization of MAPs presents a fundamental scalability bottleneck. By assigning a distinct prompt to each of the $2^n$ possible missing-view configurations for $n$ input views, the approach incurs an exponential parameter complexity of $\mathcal{O}(2^n)$, rendering it intractable for real-world applications with a large number of views. To mitigate the parameter explosion, we propose an \emph{efficient prototype tensorization} (EPT) module, which leverages atomic tensor decomposition to reduce the parameter complexity from exponential to linear in $n$, i.e., $\mathcal{O}(n)$, while preserving the expressiveness of distinct prompts.

Specifically, we define a MAPs matrix $P_M \in \mathbb{R}^{d \times 2^n}$, where $d$ is the prompt length and each column $P_M^i$ represents the prompt for the $i$-th missing-view pattern. To ensure tractability, we employ a \emph{structured parameterization} that integrates low-rank matrix factorization~\cite{sainath2013low} and \emph{tensor train} (TT) decomposition~\cite{oseledets2011tensor}, enabling efficient modeling of complex dependencies among missing-view patterns.
Specifically, we reparameterize the MAPs matrix using a factorized structure:
\begin{equation}
    P_M = A \, B^\top,
\end{equation}
where $A \in \mathbb{R}^{d \times k}$ is a shared basis matrix ($k \ll d$), and $B \in \mathbb{R}^{2^n \times k}$ contains the coefficients for each missing-view pattern. This reduces the total number of parameters from $\mathcal{O}(d \cdot 2^n)$ to $\mathcal{O}((d + 2^n) \cdot k)$. 

However, the size of $B$ still scales exponentially in $n$. To further compress $B$, we reshape it into a tensor $\mathcal{B} \in \mathbb{R}^{2 \times 2 \times \cdots \times 2 \times k}$ of order $n+1$, where each of the $n$ binary modes corresponds to the presence ($1$) or absence ($0$) of a view and the last mode indexes the k latent factors. We then apply TT decomposition to $\mathcal{B}$ which expresses it as a sequence of core tensors $\{G_1, \cdots , G_{n+1}\}$, where $G_\ell \in \mathbb{R}^{r_{\ell-1} \times 2 \times  r_\ell}$ for $\ell<=n$ and $G_{n+1} \in \mathbb{R}^{r_n \times k}$. The TT decomposition is defined as:
\begin{equation}
    \mathcal{B} (s_1, \cdots, s_n) = G_1(s_1)\cdots G_n(s_n) \cdot G_{n+1},
\end{equation}
where $s_j \in \{0, 1\}$  indicates the presence of the $j$-th view, and $G_j(s_j) \in \mathbb{R}^{r_{j-1} \times r_j}$ denotes the $s_j$-th slice of the $j$-th core tensor. The resulting coefficient vector for a missing-view pattern $m = (m_1, \cdots, m_n)$ is:
\begin{equation}
    \beta_m = G_1(m_1) \cdots G_n(m_n) \cdot G_{n+1}.
\end{equation}
Finally, the MAP for pattern $m$ is obtained as:
\begin{equation}
    P_M^m = A \, \beta_m^\top.
\end{equation}
This structured parameterization enables EPT to efficiently model complex dependencies among missing-view patterns while maintaining scalability. The total number of learnable parameters becomes $\mathcal{O}(n \cdot R^2 \cdot k + d \cdot k)$, where $R = \max(r_0, \cdots, r_n)$ is the maximum TT rank. This reduces the complexity from exponential to linear in the number of views $n$ , making EPT suitable for high-dimensional multi-view settings.

\subsection{Dynamic Contrastive Learning}
The semantic relationships among missing-view patterns in multi-view data are inherently complex, as the presence or absence of specific views critically determines their latent similarity. While patterns sharing observed views are intuitively closer in the latent space, simply counting shared views overlooks the fact that each view contributes unequally—some offer more discriminative or complementary information depending on context.

To address these challenges, we propose a \emph{dynamic contrastive learning} (DCL) strategy that explicitly models both the structural overlap and the heterogeneous importance of each view. Specifically, given $n$ input views, and each possible missing pattern can be represented as a binary vector of length $n$, resulting in $2^n$ possible patterns. We assign a learnable weight parameter $w = [w_1, w_2, ..., w_n]$ to each view. For any two missing patterns $i$ and $j$, represented by binary view-indicator vector $m_i$ and $m_j$, we compute their weighted overlap score as follows:
\begin{equation}
    s_{ij} = m_i^\top \cdot \mathrm{Sigmoid}(w) \cdot m_j
\end{equation}
where the $\mathrm{Sigmoid}$ function constrains each weight to $(0, 1)$. Based on $s_{ij}$, we construct positive pairs ($s_{ij} > 0$) that share at least one observed view, and negative pairs ($s_{ij} = 0$) with no overlap. The dynamic contrastive learning loss is designed to pull positive pairs closer in the embedding space while enforcing a margin $\alpha$ between negative pairs:
\begin{align}
    \mathcal{L}_{DCL} =\ & \frac{1}{|P|} \sum_{(i,j) \in P} \|P_M^i - P_M^j\|^2 \notag \\
    & + \frac{1}{|N|} \sum_{(i,j) \in N} \max\left(0,\, \alpha - \|P_M^i - P_M^j\|\right)^2,
\end{align}
where $P$ and $N$ denote the sets of positive and negative pairs, $\alpha$ is the distance margin, and $\|\cdot\|$ denotes the Euclidean norm.

\begin{algorithm}[t]
\caption{Training process of \emph{E2PL}}
\label{alg:E2PL_training}
\begin{algorithmic}[1]
\REQUIRE Incomplete multi-view data $(\mathcal{X}^t, \mathcal{Y}^t)$ for task $t$; view-indicator matrix $M^t$; hyperparameters $\lambda$.
\ENSURE Updated model parameters.
\FOR{each sample $i$ in $\mathcal{X}^t$}
    \STATE Impute missing views with zeros according to $m_i^t$
    \STATE Extract view-specific features $h_{i,v}^t$ for all views $v$ \hfill (Eq.7)
    \STATE Form input sequence $S_i^t = [h_{[\text{CLS}]}, h_{i,1}^t, ..., h_{i,n}^t]$ \hfill (Eq.8)
    \STATE Generate MAPs $P_M$ using EPT module
    \STATE Index the corresponding MAP $P_M^i$ with $m_i^t$ 
    \hfill (Eq.9)
    \STATE Retrieve TTP $P_T^t$ for current task
    \STATE Split $P_M^i$ into $P_M^{i,K}, P_M^{i,V}$; split $P_T^t$ into $P_T^{t,K}, P_T^{t,V}$ 
    \hfill (Eq.10)
    \STATE Inject $P_M^{i,K}, P_M^{i,V}, P_T^{t,K}, P_T^{t,V}$ into K and V vectors 
    \hfill (Eq.11)
    \STATE Forward pass to obtain representation $z_i^t$ and prediction $\hat{y}_i^t$ \hfill (Eq.12)
\ENDFOR
\STATE Compute binary cross-entropy loss $\mathcal{L}_{BCE}$ \hfill (Eq.13)
\STATE Compute dynamic contrastive learning loss $\mathcal{L}_{DCL}$ \hfill (Eq.6)
\STATE Compute total loss $\mathcal{L}_{Total} = \mathcal{L}_{BCE} + \lambda \cdot \mathcal{L}_{DCL}$ \hfill (Eq.14)
\STATE Update model parameters for current task $t$ (freeze prompts from previous tasks)
\end{algorithmic}
\end{algorithm}

\subsection{Training and Inference Pipeline}
To facilitate reproducibility and provide a clear overview of our approach, we present the complete training and inference workflows in Algorithm~\ref{alg:E2PL_training} and Algorithm~\ref{alg:E2PL_inference}, respectively. These algorithms encapsulate the core mechanisms of E2PL, including data preprocessing, prompt generation and injection, and model update, and are designed to efficiently address both class-incremental adaptation and arbitrary missing-view scenarios.

\emph{Training Pipeline.} Taking task $t$ session as an example, the training pipeline begins with the current session’s dataset $(\mathcal{X}^t, \mathcal{Y}^t)$ and the corresponding label set $\mathcal{C}^t$.
For each sample $i$, the multi-view features $(x_{i,1}^t, \cdots, x_{i,n}^t)$ are first preprocessed by imputing missing views as zero vectors, according to the view-indicator vector $m_i^t \in \{0,1\}^n$. Each available view is then projected into a shared embedding space via a view-specific linear projection encoder:
\begin{equation}
    h_{i,v}^t = \mathrm{Encoder}_v(x_{i,v}^t), \quad v=1,\cdots,n.
\end{equation}
The input sequence to the task $t$ forward module, which is composed of $N$ stacked Transformer layers, is formed by concatenating the [CLS] token embedding with the embeddings of all views, yielding
\begin{equation}
    S_i^t = [h_{\mathrm{[CLS]}}, h_{i,1}^t, \cdots, h_{i,n}^t].
\end{equation}
To explicitly encode the missing-view pattern, we first use the EPT module to generate MAPs $P_M = \{P_M^1, \cdots, P_M^{2^n}\}$.
For each sample $i$, the view-indicator vector $m_i^t$ is used to index the corresponding missing-aware prompt $P_M^i$ from $P_M$:
\begin{equation}
    P_M^i = P_M[m_i^t].
\end{equation}
Meanwhile, for the current task $t$, the corresponding task-tailored prompt $P_T^t$ is retrieved. Both $P_M^i$ and $P_T^t$ are split equally along the feature dimension into key and value parts:
\begin{equation}
    P_M^{i,K}, P_M^{i,V} = \mathrm{Split}(P_M^i),   \quad
    P_T^{t,K}, P_T^{t,V} = \mathrm{Split}(P_T^t)
\end{equation}
At each transformer layer $l$, $P_M^{i,K} $ and $P_M^{i,V}$ are added to the key and value representations, respectively:
\begin{equation}
    K^{(l)} = K^{(l)}_{\text{orig}} + P_M^{i,K} + P_T^{t,K}, \quad
    V^{(l)} = V^{(l)}_{\text{orig}} + P_M^{i,V} + P_T^{t,V}
\end{equation}
where $K^{(l)}_{\text{orig}}$ and $V^{(l)}_{\text{orig}}$ denote the original key and value matrices computed via $\mathrm{QKV}$ projection, respectively.
The prompt-augmented sequence is then processed through multi-head attention and MLP layers.
Finally, the [CLS] token output from the last layer, denoted as $z_i^t$ used as the global representation of the multi-view sample. This representation is then passed to a task-specific classification head to predict the current class space $\mathcal{C}^{t}$:
\begin{equation}
    \hat{y}_i^t = \sigma(W_c^t z_i^t + b_c^t),
\end{equation}
where $\sigma(\cdot)$ is the sigmoid activation function, and $W_c^t$, $b_c^t$ are the parameters of the classification head.
\begin{algorithm}[t]
\caption{Inference process of \emph{E2PL}}
\label{alg:E2PL_inference}
\begin{algorithmic}[1]
\REQUIRE Test sample $\mathcal{X}_i$; view-indicator vector $m_i$.
\ENSURE Final prediction $\hat{Y}_i$ over all classes.
\STATE Impute missing views with zeros according to $m_i$
\STATE Extract view-specific features and form input sequence $S_i$
\STATE Generate MAP $P_M^i$ using EPT module with $m_i$
\FOR{each task $\tau = 1$ to $T$}
    \STATE Load task-tailored prompt $P_T^\tau$
    \STATE Split $P_M^i$ into $P_M^{i,K}$ and $P_M^{i,V}$; split $P_T^\tau$ into $P_T^{\tau,K}$ and $P_T^{\tau,V}$
    \STATE Inject $P_M^{i,K}, P_M^{i,V}, P_T^{\tau,K}, P_T^{\tau,V}$ into K and V vectors
    \STATE Forward pass to get task-specific logit $\hat{y}_i^\tau$  
\ENDFOR
\STATE Concatenate all logits: $\hat{Y}_i = [\hat{y}_i^1; \hat{y}_i^2; ...; \hat{y}_i^T]$ \hfill (Eq.15)
\end{algorithmic}
\end{algorithm}

\emph{Objective Function.}
To simultaneously enhance predictive robustness and representational capacity, we formulate a unified objective function that synergistically combines two distinct loss terms. The first is the standard binary cross-entropy (BCE) loss, which governs performance on the primary classification task, i.e.,
\begin{equation}
\mathcal{L}_{BCE} = -\frac{1}{|\mathcal{Y}^t|} \sum_{i \in \mathcal{Y}^t} \left[ y_i^t \log \hat{y}_i^t + (1 - y_i^t) \log (1 - \hat{y}_i^t) \right],
\end{equation}
where $y_i^t$ is the ground truth binary label and $\hat{y}_i^t$ is the corresponding model prediction.
The second, a dynamic contrastive learning loss $\mathcal{L}_{DCL}$, regularizes the latent space by enforcing semantic consistency and discriminability among diverse missing-view patterns.

Hence, the final objective for \textsf{E2PL}, denoted $\mathcal{L}_{Total}$, is the weighted aggregation of these two components:
\begin{equation}
\mathcal{L}_{Total} =
\mathcal{L}_{BCE} + \lambda \cdot \mathcal{L}_{DCL}, 
\end{equation}
where $\lambda$ is a hyperparameter that balances two loss terms.

\emph{Inference Pipeline.}
During the inference stage, for a test sample $\mathcal{X}_i$, the MAP $P_M^i$ is generated by the EPT module according to the view-indicator vector $m_i$. 
Meanwhile, all TTPs from the $T$ tasks are attached in parallel to the backbone network, resulting in $T$ forward propagation pathways, as shown in Figure~\ref{fig:architecture} (b).
The [CLS] representation from each pathway is fed into its corresponding classification head to obtain the task-specific logit. All logits are then concatenated to form the final prediction vector:
\begin{equation}
    \hat{Y} = [\hat{y}^1; \cdots; \hat{y}^T],
\end{equation}
where $[\cdot;\cdot]$ denotes the concatenation operation. The final multi-label prediction is obtained by independently thresholding each sigmoid-normalized logit, which is the standard practice in multi-label and binary classification tasks.

%% file: 4.Experiment.tex
\section{Experiment}
\begin{table*}[ht]
\centering

\setlength{\tabcolsep}{4.5pt}
\caption{Evaluation on three datasets at $R_V = 30\%$. ``Memory'' denotes the number of replay samples; ``-R'' indicates the memory allocated to the method. \textbf{Bold} and \underline{underlined} values indicate the \textbf{best} and \underline{second-best} results, respectively.}
\begin{tabular}{l c c cccc cccc cccc}
\toprule
\multirow{3}{*}{\textbf{Methods}} & \multirow{3}{*}{\textbf{Types}} & \multirow{3}{*}{\textbf{Memory}}
& \multicolumn{4}{c}{\emph{ESPGame}} 
& \multicolumn{4}{c}{\emph{IAPRTC12}}
& \multicolumn{4}{c}{\emph{MIRFLICKR}}\\
\cmidrule(lr){4-7} \cmidrule(lr){8-11} \cmidrule(lr){12-15}
& & &\multicolumn{1}{c}{Avg.} & \multicolumn{3}{c}{Last} 
 &\multicolumn{1}{c}{Avg.} & \multicolumn{3}{c}{Last}
 &\multicolumn{1}{c}{Avg.} & \multicolumn{3}{c}{Last}\\
\cmidrule(lr){4-4} \cmidrule(lr){5-7} \cmidrule(lr){8-8} \cmidrule(lr){9-11} \cmidrule(lr){12-12} \cmidrule(lr){13-15}
& & & mAP & CF1 & OF1 & mAP & mAP & CF1 & OF1 & mAP & mAP & CF1 & OF1 & mAP  \\
\midrule
Upper-bound & Baseline & 0 & - & 2.99 & 13.54 & 30.17 & - & 3.24 & 17.49 & 32.45 & - & 36.71 & 51.25 & 62.47 \\
\midrule
DICNet & IMvMLC & \multirow{8}{*}{0} & 20.01 & 0.61 & 5.82 & 17.82 & 23.42 & 0.58 &  12.67 &  22.50 &  61.29 & 19.04 & 43.13 & 49.68 \\
MTD & IMvMLC & & 22.89 & 0.95 & 8.92 & 19.79 & 26.98 & 1.23 & 13.30 & 25.12 & 63.00 & 22.34 & 43.66 & 52.11 \\
AIMNet &IMvMLC&  & 24.78 & 1.42 & 8.82 & 22.59 & 28.17 & 2.10 & 13.34 & 27.57 & 63.60 & 25.79 & 45.42 & 53.28 \\
TSIEN &IMvMLC &  & 20.85 & 0.73 & 6.17 & 19.63 & 22.50 & 0.92 & 10.35 & 20.47 & 61.92 & 12.95 & 33.89 & 52.94 \\
SIP & IMvMLC&  & 24.79 & 1.02 & 8.11 & 23.21 & 24.23 & 2.18 & 13.21 & 20.15 & 61.93 & 24.92 & 38.13 & 49.92 \\
KRT &MLCIL &  & 21.11 & 0.72 & 5.23 & 19.97 & 26.78 & 1.45 & 11.21 & 25.44 & 57.07 & 10.63 & 35.12 & 50.59 \\
MULTI-LANE & MLCIL&  & 22.85 & 0.97 & 7.14 & 20.88 & \underline{28.41} & 2.24 & 13.61 & 27.91 & 58.98 & 11.61 & 37.09 & 52.13 \\
\midrule
AIMNet-R &IMvMLC &\multirow{4}{*}{200} & 25.59 & 1.47 & 8.82 & 23.33 & 28.40 & \underline{2.53} & \underline{13.37} & 28.34 & \underline{64.59} & 26.43 & 41.36 & 54.18 \\
SIP-R & IMvMLC&   & \underline{25.89} & 1.53 & 8.81 & 24.13 & 26.61 & 1.92 & 11.87 & 25.29 & 63.98 & 26.23 & 40.11 & 52.94 \\
PRS & MLCIL & & 20.12 & 0.58 & 5.82 & 17.23 & 22.34 & 1.13 & 9.24 & 22.11 & 56.66 & 9.67 & 33.13 & 51.88 \\
KRT-R & MLCIL& & 22.56 & 0.79 & 7.21 & 21.12 & 27.42 & 1.69 & 12.19 & 26.77 & 61.01 & 16.12 & 39.98 & 52.62 \\
\midrule
AIMNet-R &IMvMLC &\multirow{4}{*}{2/class} & 25.79 & \underline{1.56} & \underline{9.84} & 24.01 & 28.03 & 2.12 & 12.34 & 27.56 & 64.29 & \underline{26.45} & \underline{46.32} & \underline{54.30} \\
SIP-R & IMvMLC&   & 25.36 & 1.47 & 8.97 & \underline{24.43} & 26.72 & 1.83 & 11.13 & 25.34 & 63.71 & 25.80 & 39.50 & 52.63 \\
PRS & MLCIL & & 20.38 & 0.62 & 6.42 & 17.83 & 22.44 & 1.18 & 9.55 & 21.44& 56.65 & 9.28 & 31.98 & 50.23 \\
KRT-R & MLCIL& & 22.89 & 0.82 & 7.56 & 21.44 & 28.01 & 1.78 & 12.56 & 26.72& 60.11 & 15.34 & 39.50 & 52.77 \\
\midrule
\textsf{E2PL} & IMvMLCIL &  0 & \textbf{28.29} & \textbf{2.37} & \textbf{11.37} & \textbf{27.10} & \textbf{30.74} & \textbf{2.91}& \textbf{14.82}& \textbf{30.40} & \textbf{66.76} & \textbf{30.73} & \textbf{47.14} & \textbf{59.08} \\
\bottomrule
\end{tabular}

\label{tab:sota}
\end{table*}


\emph{Datasets.} 
Following the protocol established in previous works \cite{liu2024attention}, we conduct experiments on three widely used multi-view multi-label datasets: \emph{ESPGame} \cite{von2004labeling}, \emph{IAPRTC12} \cite{grubinger2006iapr}, and \emph{MIRFLICKR} \cite{huiskes2008mir}. Each dataset is represented by six distinct feature types, including GIST, HSV, DenseHue, DenseSift, RGB, and LAB. Detailed statistics for all datasets are provided in \textbf{Appendix A}. For each dataset, we randomly choose 15\% samples for the test, 15\% for validation, and the rest for training.

\emph{Metrics.} 
Following \cite{dong2023knowledge} and \cite{zhang2025l3a}, we adopt the mean average precision (mAP) as the primary evaluation metric for each incremental task, and report both the average mAP (the average of the mAP across all tasks) and the last mAP (the mAP of the
last task). We also report the per-class F1 score (CF1) and
overall F1 score (OF1) for performance evaluation.

\emph{Baselines.} 
We compare \textsf{E2PL} with five state-of-the-art IMvMLC methods: DICNet~\cite{liu2023dicnet}, MTD~\cite{liu2023masked}, AIMNet~\cite{liu2024attention}, TSIEN~\cite{tan2024two}, and SIP~\cite{liu2024partial};
and three MLCIL methods: PSR~\cite{kim2020imbalanced}, KRT~\cite{dong2023knowledge}, and MULTI-LANE~\cite{de2024less}. We include the Upper-bound baseline, which assumes access to all task data simultaneously and represents the ideal performance.
Following \cite{dong2023knowledge}, we set three different memory sizes: 0, 200, and 2/class.
To ensure fair and rigorous experimental comparisons, we strictly adhere to the original implementation protocols for all MLCIL methods. As these methods are limited to single-view inputs, we independently train each method on every available view and report the best performance achieved across all views.

\emph{Implementation Details.}
Since all datasets are originally complete in terms of view availability, we artificially introduce missing data to emulate real-world incomplete scenarios. 
Following~\cite{liu2023dicnet}, we generate a missing instance rate of $p\%$ (i.e., view missing rate $R_V = p\%$), by randomly removing $p\%$ of the instances from each view, while ensuring every instance remains observable in at least one view. To simulate a class-incremental learning scenario, each dataset is partitioned into $T=7$ incremental tasks, with an equal number of classes introduced at each stage. For our proposed method, the dimension $d$ of all prompts is set to 128, and all TT-ranks are set as $r_0 = r_1 = \cdots = r_n = 2$. Other hyperparameters are set as follows: $N=3$, $k=4$, $\alpha=1$, and $\lambda=0.001$. The Adam optimizer is used for all experiments, with a batch size of 128 and a learning rate of 0.02. The detailed experimental setups are provided in \textbf{Appendix B}.
For fairness, all results are averaged over five times.

 \begin{figure*}[t]
  \includegraphics[width=\textwidth]{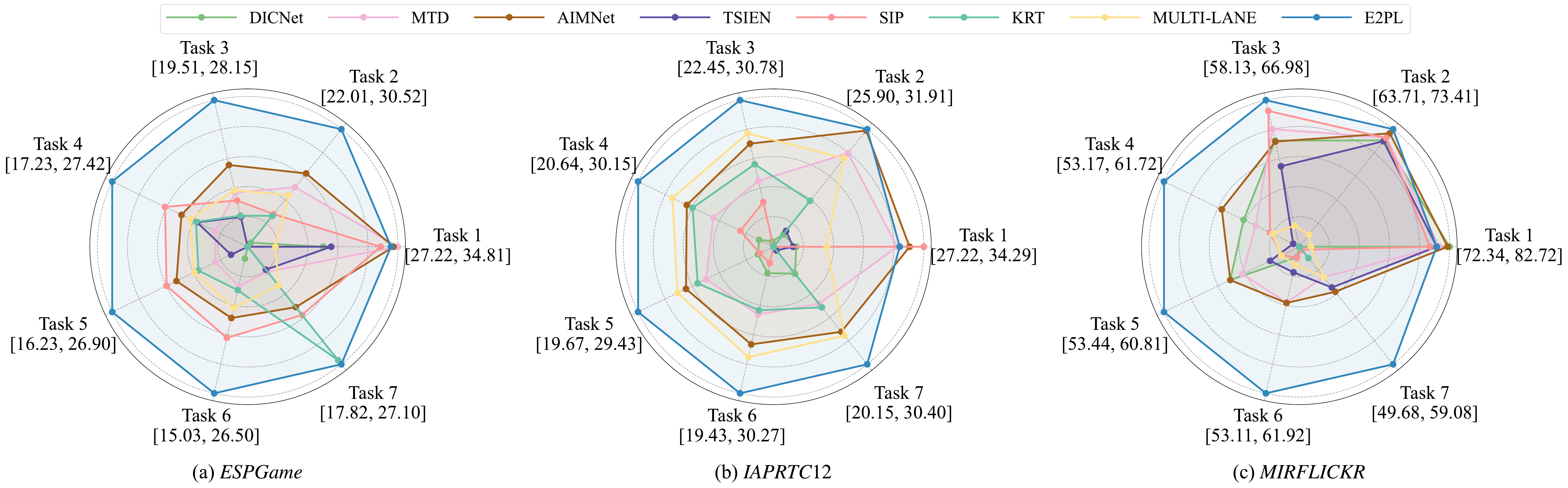}
  \caption{Comparison of performance (mAP) on three datasets at $R_V = 30\%$ over $T=7$ incremental tasks (Memory = 0).}
  \label{fig:incremental}
  \vspace*{-0.15in}
\end{figure*}

\subsection{Comparison Study}
Table~\ref{tab:sota} summarizes the performance across all datasets, showing that \textsf{E2PL} consistently outperforms all baselines.
On the \emph{MIRFLICKR} dataset, under the strict zero-memory setting, \textsf{E2PL} achieves a mAP of 66.76, surpassing the best IMvMLC baseline AIMNet-R and the leading MLCIL method KRT-R by 2.47 and 5.75 points, corresponding to relative gains of 3.8\% and 9.4\%.
For the last-increment mAP, \textsf{E2PL} reaches 59.08, exceeding all baselines by 4.78 to 6.46 points. Similarly, \textsf{E2PL} achieves the highest CF1 and OF1 scores without memory replay. Consistent trends are observed on \emph{ESPGame} and \emph{IAPRTC12}, where \textsf{E2PL} achieves the best results on all metrics. Notably, \textsf{E2PL} is the only method that consistently maintains high performance under the most challenging incremental learning scenarios, underscoring its robustness and generalizability in both class-incremental and incomplete multi-view settings. 

Figure~\ref{fig:incremental} provides a fine-grained depiction of mAP progression across incremental learning tasks. Unlike other methods that suffer significant drops as new classes are introduced, \textsf{E2PL} maintains stable performance, with fluctuations below 2\% across all increments. This resilience to forgetting is attributed to its prompt-based architecture, which preserves previously learned knowledge while effectively integrating new information, ensuring both stability and adaptability in class incremental learning.

\begin{figure}[t]
\centering
  \includegraphics[width=1\columnwidth]{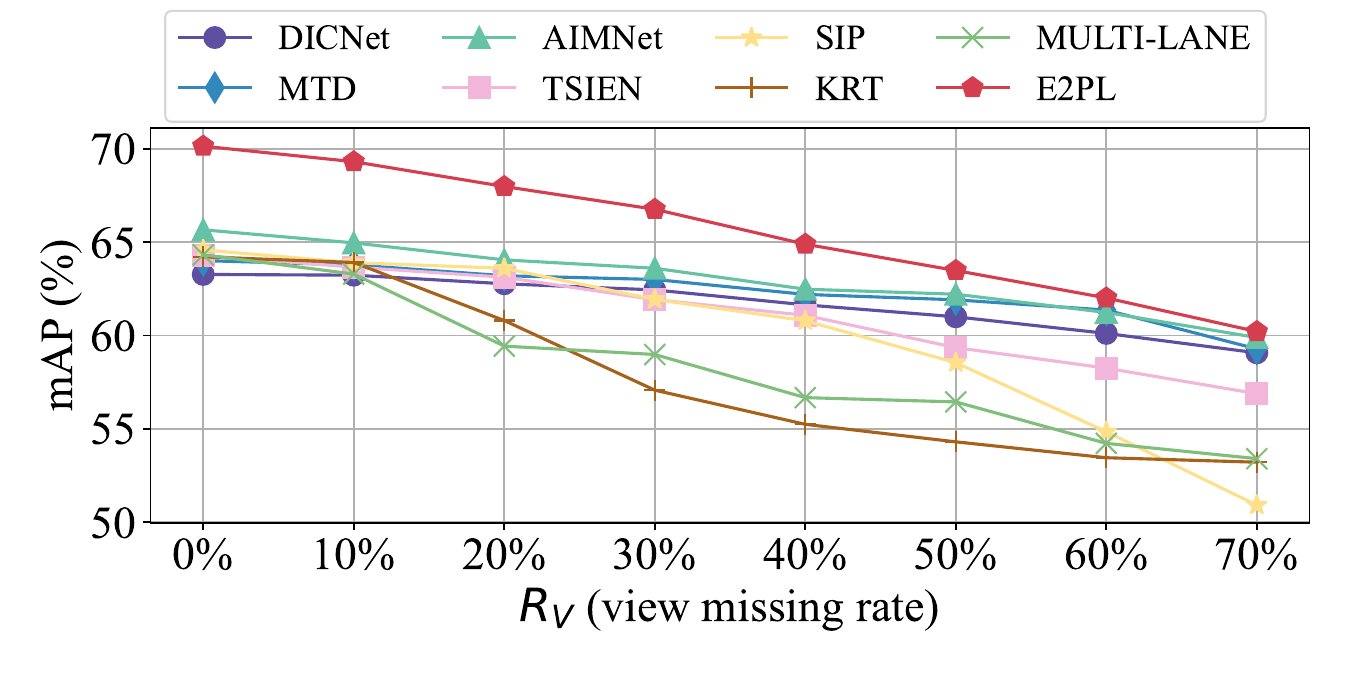}
  \caption{Performance demonstration under different view missing rates on the \emph{MIRFLICKR} dataset.}
  \label{fig:mr}
  \vspace*{-0.15in}
\end{figure}

\subsection{Effect of View Missing Rates}
Figure~\ref{fig:mr} illustrates the robustness of different methods under varying view missing rates ($R_V$) in the CIL setting. \textsf{E2PL} consistently achieves the highest mAP across all missing rates. At $R_V=0\%$, \textsf{E2PL} attains a mAP of 70.14, exceeding the best baseline AIMNet by 4.48 points. When $R_V$ rises to $30\%$, \textsf{E2PL} maintains a leading mAP of 66.76, while AIMNet drops to 63.60, resulting in a margin of 3.16 points. Even at $R_V=70\%$, \textsf{E2PL} achieves 60.21 mAP, outperforming the nearest competitor AIMNet by 6.13 points. Notably, as $R_V$ increases, the performance gap between \textsf{E2PL} and the baseline methods gradually narrows, decreasing from 4.48\% at $R_V=0\%$ to 2.0\% at $R_V=70\%$-which is anticipated, as higher missing rates mean less available information and greater learning difficulty, particularly in class incremental settings. Nevertheless, the prompt-based architecture of \textsf{E2PL} exhibits substantial robustness and adaptability, consistently outperforming state-of-the-art methods and ensuring reliable performance even under severe view missing scenarios.
Additional experiments on two other datasets, detailed in \textbf{Appendix C}, further confirm the effectiveness and generalizability of \textsf{E2PL}.


\subsection{Analysis of Effectiveness and Efficiency}
To rigorously assess the effectiveness and efficiency of \textsf{E2PL}, we perform controlled experiments by substituting the EPT module with MAP~\cite{lee2023multimodal}, MSP~\cite{jang2024towards}, and EPE-P~\cite{chen2024epe} within a unified framework.
Here, the parameter count specifically refers to the number of prompt parameters required to model all possible missing-view patterns, where $n$  is the total number of views.
Other variables, including $d$, $r$, $R$, and $k$, are treated as constants, and all other factors are held fixed. As summarized in Table~\ref{tab:Eff}, MAP and MSP exhibit exponential parameter growth ($\mathcal{O}(2^n)$), and EPE-P shows cubic complexity, while EPT achieves linear growth ($\mathcal{O}(n)$) owing to its efficient modeling strategy.
EPT leverages atomic tensor decomposition for dimension reduction, enabling customized modeling for each missing-view pattern while simultaneously enhancing interactions and knowledge sharing across different patterns. Notably, EPT achieves the highest mAP among all compared methods, with minimal storage overhead. These results convincingly demonstrate both the effectiveness and efficiency of EPT in the IMvMLCIL setting. We further validate the superior efficiency of \textsf{E2PL} through experiments on two additional datasets in \textbf{Appendix D}.

Table~\ref{tab:Eff3} further elucidates the resource efficiency of our approach. \textsf{E2PL} requires only 0.026M parameters—an 86\% reduction compared to MAP (0.193M)—and maintains a competitive training time of 335~ms per epoch. The marginally increased training time, relative to MAP (268~ms), is primarily due to the dynamic generation of all missing-aware prompts during optimization. In the inference phase, however, \textsf{E2PL} utilizes the missing-view indicator to generate only the necessary prompt for each instance, yielding a significant advantage in inference speed (11.4~ms) over MSP (27.6~ms) and EPE-P (30.1~ms), and closely approaching MAP (10.5~ms). GPU memory consumption remains moderate at 889~MB, further highlighting the method’s practical efficiency. \textsf{E2PL} thus establishes a new paradigm for IMvMLCIL by simultaneously advancing both effectiveness and efficiency in large-scale, real-world scenarios.

\begin{table}[t]
\centering
\setlength{\tabcolsep}{5.5pt}
\caption{Analysis of effectiveness and efficiency on the \emph{MIRFLICKR} dataset at $R_V = 30\%$. \textbf{Bold} and \underline{underlined} values indicate the \textbf{best} and \underline{second-best} results, respectively.}
\begin{tabular}{lccc }
\toprule
\textbf{Methods} & \textbf{\# Params} & \textbf{Complexity} & \textbf{mAP}  \\
\midrule
MAP & $(2^{n})\cdot d $ & $\mathcal{O}(2^{n})$ &65.10$\pm$0.11 \\
MSP & $n \cdot d $ & $\mathcal{O}(n)$&64.66$\pm$0.09 \\
EPE-P & $n^3 + d \cdot r $ & $\mathcal{O}(n^3)$&\underline{65.57$\pm$0.05} \\
EPT & $n \cdot R^2 \cdot k + d \cdot k $ & $\mathcal{O}(n)$& \textbf{66.76$\pm$0.03} \\
\bottomrule

\end{tabular}
\label{tab:Eff}
\end{table}




\subsection{Ablation Study}

To assess the effectiveness of each proposed component, we conduct an ablation study on the \emph{MIRFLICKR} dataset, as shown in Table~\ref{tab:Ablation}. Using only task-tailored prompts (TTPs) yields an mAP of 62.04, serving as the baseline for class-incremental adaptation. Introducing missing-aware prompts (MAPs) increases the mAP to 64.18, underscoring the value of explicitly modeling incomplete view configurations. Incorporating efficient prototype tensorization (EPT) further improves the mAP to 65.71; importantly, EPT not only reduces parameter complexity but also enhances interaction modeling across different missing-view scenarios, enabling more expressive and scalable prompt generation. The addition of dynamic contrastive learning (DCL) results in a further performance gain (mAP 65.38), as DCL dynamically captures the underlying relationships among various missing patterns, facilitating robust representation learning. When all modules are integrated, \textsf{E2PL} achieves the highest performance, outperforming all partial configurations. These results clearly demonstrate that each module addresses a distinct aspect of the IMvMLCIL task, and their synergistic integration is key for achieving state-of-the-art performance.
Furthermore, the effectiveness of each module in \textsf{E2PL} is substantiated by consistent ablation studies on two additional datasets, with detailed results in \textbf{Appendix E}.

\begin{table}[t]
    \centering
    \caption{Resource consumption and computational efficiency of different methods. \textbf{Bold} and \underline{underlined} values indicate the \textbf{best} and \underline{second-best} results, respectively. TT/E refers to the training time per epoch for the model, IT/B denotes the inference time for a single batch sample, and MGM indicates the maximum GPU memory usage recorded during the experimental process.}
    \setlength{\tabcolsep}{3pt}
    \label{tab:times}
    \begin{tabular}{l c c c c}
        \toprule
        \textbf{Methods} & \textbf{Params (M)}  & \textbf{TT/E (ms)} & \textbf{IT/B (ms)} & \textbf{MGM (MB)} \\
        \midrule
        MAP & 0.193 & \textbf{268} & \textbf{10.5} &  998 \\
        MSP & 0.049& \underline{315} & 27.6 & \underline{822} \\
        EPE-P& \underline{0.044}& 405 & 30.1 & \textbf{801}  \\
        \textsf{E2PL} &\textbf{0.026} &335 & \underline{11.4} & 889\\
        \bottomrule
    \end{tabular}
    \label{tab:Eff3}
\end{table}

\begin{table}[t]
\centering
\setlength{\tabcolsep}{8.3pt}
\newcommand{\cmark}{\checkmark}
\caption{Ablation results of \textsf{E2PT} on the \emph{MIRFLICKR} dataset at $R_V = 30\%$. \textbf{Bold} and \underline{underlined} values indicate the \textbf{best} and \underline{second-best} results, respectively.}
\begin{tabular}{ccccc}
\toprule
\multicolumn{4}{c}{\textbf{Components}} & \multirow{2}{*}{\textbf{mAP}}\\
\cmidrule(lr){1-4} 
\textbf{+TTPs} & \textbf{+MAPs} & \textbf{+EPT} & \textbf{+DCL} & \\
\midrule
      & \cmark  & \cmark & \cmark & 62.04$\pm$0.11 \\
\cmark &         &   &   & 64.18$\pm$0.05\\
\cmark      & \cmark  &        & \cmark & \underline{65.71$\pm$0.08}\\
\cmark &  \cmark  & \cmark &  &  65.38$\pm$0.07\\
\cmark &  \cmark  & \cmark & \cmark &  \textbf{66.76$\pm$0.03}\\
\bottomrule
\end{tabular}
\label{tab:Ablation}
\vspace*{-0.15in}
\end{table}

%% file: 5.Conclusion.tex
\section{Conclusion}
\label{sec:Conclusions}

In this work, we propose \textsf{E2PL}, an effective and efficient prompt learning framework that integrates TTPs for class-incremental adaptation with MSPs for flexible handling of missing views.
Our EPT module reduces prompt space complexity from exponential to linear and enables expressive modeling across diverse missing-view scenarios.
Additionally, the DCL strategy further enhances robustness by capturing latent relationships among missing patterns.
Extensive experiments on three benchmarks demonstrate that \textsf{E2PL} consistently outperforms existing methods, highlighting its effectiveness and scalability in incremental scenarios.


\section*{Acknowledgments}
This work is supported by the Leading Goose R\&D Program of Zhejiang under Grant No. 2024C01109, the Zhejiang Provincial Natural Science Foundation of China under Grant No.QN26F020098, the Young Doctoral Innovation Research Program of Ningbo Natural Science Foundation Grant No. 2024J207, and the Ningbo Yongjiang Talent Program Grant 2024A-158-G.
Yangyang Wu is the corresponding author of the work.

%% file: 6.Appendix.tex
\section{Dataset Statistics and Descriptions}
For our experiments, we selected three popular benchmark datasets: \emph{ESPGame} \cite{von2004labeling}, \emph{IAPRTC12} \cite{grubinger2006iapr}, and \emph{MIRFLICKR} \cite{huiskes2008mir}.
All datasets consist of six distinct views, including GIST, HSV, Hue,
Sift, RGB, and LAB. Table \ref{tab:datasets} presents an overview of the all datasets, including the number of samples (\# Samples), the number of labels (\# Labels), and average labels per sample (\# Label Density). The details of each dataset are described as follows:
\begin{itemize}
    \item \emph{ESPGame}: The \emph{ESPGame} dataset consists of 20770 images and 20 categories, and each image can be associated
with multiple tags.
    \item \emph{IAPRTC12}: The \emph{IAPRTC12} dataset contains 20000 images from the IAPR project, annotated with multiple categories such as indoor and outdoor scenes, objects, and animals. In our experiments, we choose 19627 samples with 291 categories annotated.
    \item \emph{MIRFLICKR}: The \emph{MIRFLICKR} dataset includes 25000 images from the Flickr platform and includes 38 tags.
\end{itemize}

\begin{table}[t]
    \centering
    \caption{Statistics of all datasets used in our experiments.}
    \begin{tabular}{lccccc}
        \toprule
        Datasets & \# Samples & \# Labels & \# Label Density \\
        \midrule
        \textit{ESPGame}   & 20770 & 268 & 4.686 \\
        \textit{IAPRTC12}  & 19627 & 291 & 5.719 \\
        \textit{MIRFLICKR} & 25000 & 38  & 4.716 \\
        \bottomrule
    \end{tabular}
    \label{tab:datasets}
\end{table}

\begin{table}[h]
    \centering
    \setlength{\tabcolsep}{1pt}
    \caption{Analysis of efficiency on the \emph{ESPGame} and \emph{IAPRTC12} datasets at $R_V = 30\%$. \textbf{Bold} and \underline{underlined} values indicate the \textbf{best} and \underline{second-best} results, respectively.}
    \begin{tabular}{l c c c c}
        \toprule
        \multirow{2}{*}{\textbf{Methods}} & \multirow{2}{*}{\textbf{\# Params}} & \multirow{2}{*}{\textbf{Complexity}} & \multicolumn{2}{c}{\textbf{mAP}} \\
        \cmidrule(lr){4-5}
        & & & \emph{ESPGame} & \emph{IAPRTC12} \\
        \midrule
        MAP & $(2^n) \cdot d$ & $\mathcal{O}(2^n)$ & 27.12 $\pm$ 0.08 & 29.78 $\pm$ 0.02 \\
        MSP & $n \cdot d$ & $\mathcal{O}(n)$ & 26.78 $\pm$ 0.01 & 29.19 $\pm$ 0.04 \\
        EPE-P & $n^3 + d \cdot r$ & $\mathcal{O}(n^3)$ & \underline{27.44 $\pm$ 0.03} & \underline{29.90 $\pm$ 0.04} \\
        EPT & $n \cdot R^2 \cdot k + d \cdot k$ & $\mathcal{O}(n)$ & \textbf{28.29 $\pm$ 0.02} & \textbf{30.74 $\pm$ 0.04} \\
        \bottomrule
    \end{tabular}

        \label{tab:Eff2}
\end{table}

\section{Experimental Setup}
The experiments were conducted on the Ubuntu 22.04 system with Intel(R) Xeon(R) Gold 6326 CPU @ 2.90GHz, 256GB RAM, and NVIDIA L40 GPU (46GB memory size). 

\emph{Class Incremental Task Setup.}
We evaluate our method under the \emph{incomplete multi-view multi-label class incremental learning} (IMvMLCIL) setting, as defined above. The dataset is organized into $T$ incremental sessions $\{ (\mathcal{X}^1, \mathcal{Y}^1, \mathcal{C}^1), \cdots, (\mathcal{X}^T, \mathcal{Y}^T, \mathcal{C}^T) \}$, where each session $t$ introduces a disjoint class set $\mathcal{C}^t$ ($\mathcal{C}^m \cap \mathcal{C}^n = \emptyset$ for $m \neq n$). Given $C$ total classes, the first session contains $C_{\text{base}}$ base classes, and the remaining $C - C_{\text{base}}$ classes are evenly divided among the subsequent $T-1$ sessions. The number of new classes per incremental session is
\begin{equation}
    C_{\text{inc}} = \frac{C - C_{\text{base}}}{T - 1},
\end{equation}
where $C_{\text{inc}}$ is required to be an integer for balanced partitioning.

In the multi-label scenario, each sample may be annotated with multiple labels spanning different sessions. To ensure all relevant supervisory signals are available, a sample $x_i$ is included in session $t$ if its label set overlaps with $\mathcal{C}^t$, i.e.,
\begin{equation}
    x_i \in \mathcal{X}^t \iff \text{Label}(x_i) \cap \mathcal{C}^t \neq \emptyset.
\end{equation}
This protocol allows the model to observe samples containing both current and previously seen classes, which is essential for mitigating catastrophic forgetting in MvMLCIL. Notably, this may result in the same sample appearing in multiple sessions—a natural property of multi-label incremental learning.

For all datasets, we set $T = 7$ and adopt the following configurations: \emph{ESPGame} ($C_{\text{base}} = 208$, $C_{\text{inc}} = 10$), \emph{IAPRTC12} ($C_{\text{base}} = 231$, $C_{\text{inc}} = 10$), and \emph{MIRFLICKR} ($C_{\text{base}} = 7$, $C_{\text{inc}} = 8$). This setup enables a rigorous evaluation of model plasticity and stability under realistic, complex IMvMLCIL conditions.

\section{Effect of View Missing Rates}
\begin{figure}[h]
\centering
  \includegraphics[width=\columnwidth]{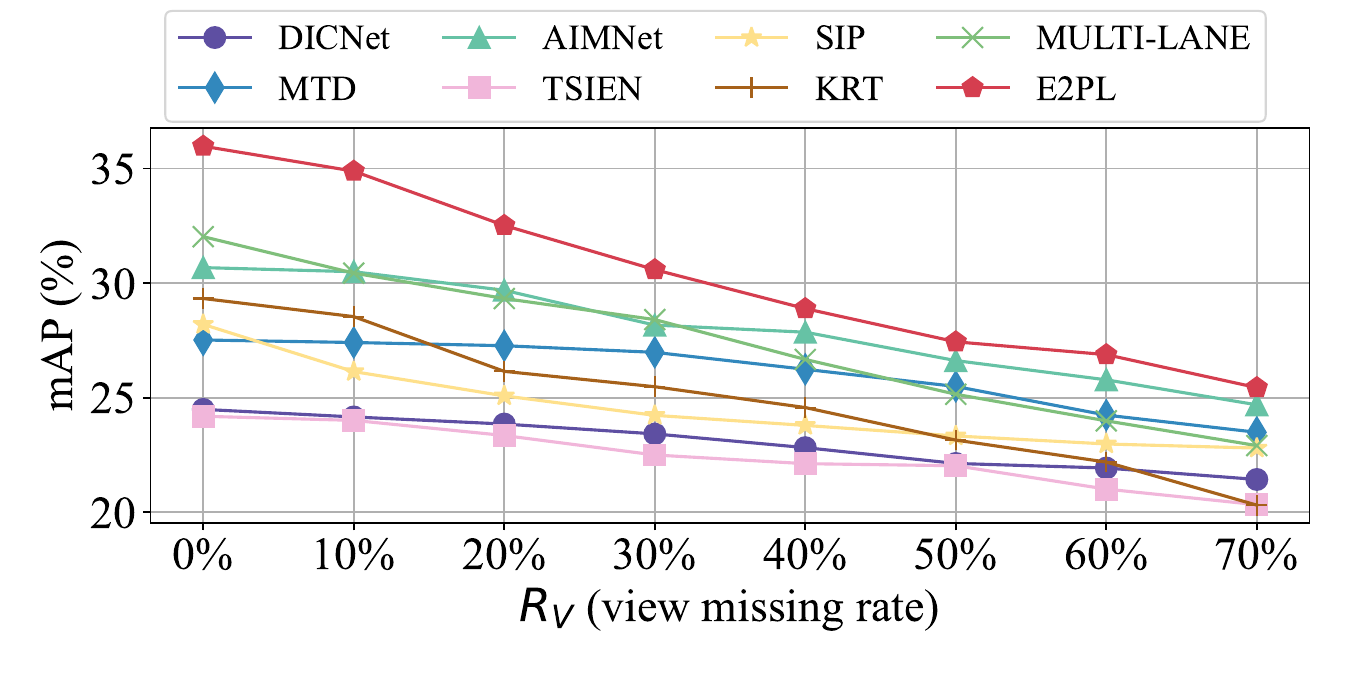}
  \caption{Performance demonstration under different view missing rates on the \emph{IAPRTC12} dataset.}
  \label{fig:mr3}
\end{figure}

\begin{figure}[h]
\centering
  \includegraphics[width=\columnwidth]{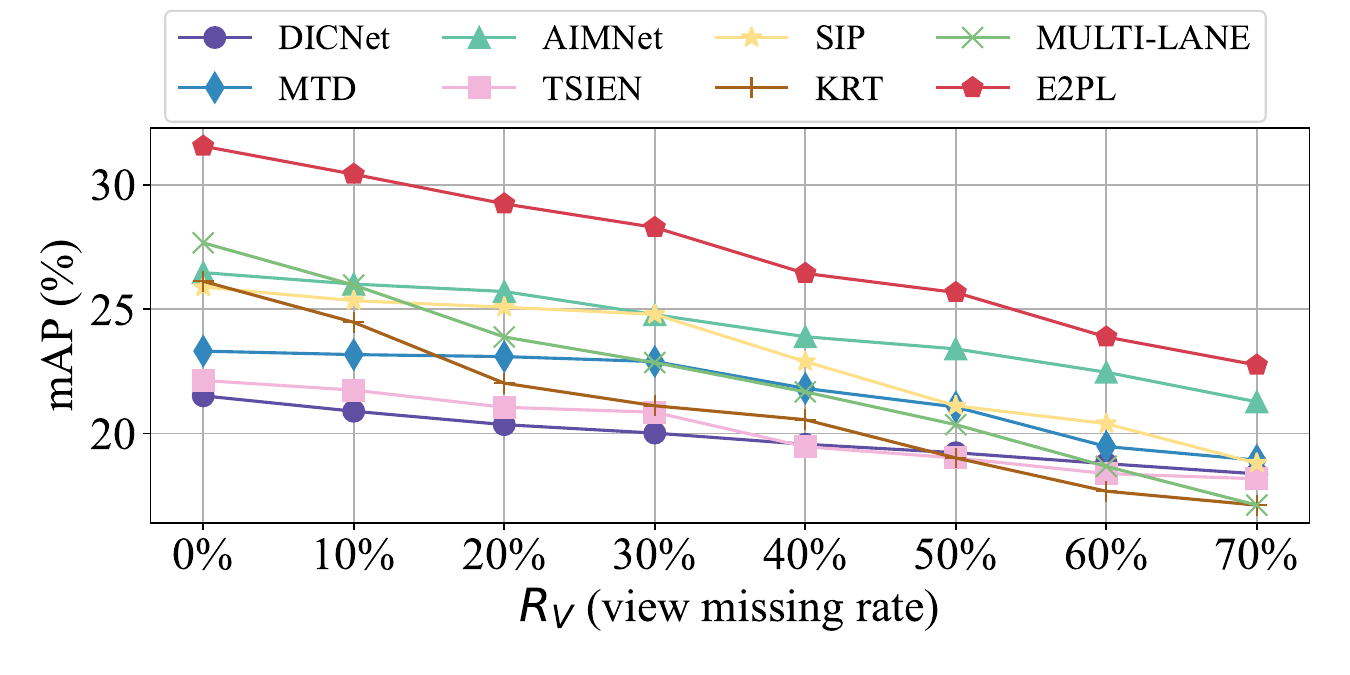}
  \caption{Performance demonstration under different view missing rates on the \emph{ESPGame} dataset.}
  \label{fig:mr2}
\end{figure}

To further validate the robustness of \textsf{E2PL} under varying degrees of view incompleteness, we conduct additional experiments on the \emph{ESPGame} and \emph{IAPRTC12} datasets, systematically increasing the view missing rate $R_V$ from 0\% to 70\%. Figure~\ref{fig:mr2} and Figure~\ref{fig:mr3} present the mean Average Precision (mAP) of \textsf{E2PL} and several state-of-the-art baselines under different $R_V$ settings.

On the \emph{ESPGame} dataset, \textsf{\textsf{E2PL}} consistently achieves the highest mAP across all missing rates. For example, at $R_V=0\%$, \textsf{\textsf{E2PL}} obtains an mAP of 31.56, outperforming the best baseline (MULTI-LANE, 27.67) by 3.89 points. As the missing rate increases to 30\%, \textsf{\textsf{E2PL}} maintains a leading mAP of 28.29, which is 5.44 points higher than both AIMNet and MULTI-LANE (22.85). Even under the most challenging scenario with $R_V=70\%$, \textsf{\textsf{E2PL}} achieves 22.75 mAP, still surpassing AIMNet (21.27) by 1.48 points. A similar trend is observed on the \emph{IAPRTC12} dataset. \textsf{\textsf{E2PL}} achieves an mAP of 35.98 at $R_V=0\%$, which is 3.95 points higher than MULTI-LANE (32.03). As $R_V$ increases to 30\%, \textsf{\textsf{E2PL}} maintains a strong lead with 30.74 mAP, compared to 28.41 for MULTI-LANE and 28.17 for AIMNet. Even at $R_V=70\%$, \textsf{\textsf{E2PL}} achieves 25.44 mAP, outperforming all baselines by at least 2.53 points.

These findings are highly consistent with the results reported in the main text (see Section 5.2), where \textsf{\textsf{E2PL}} demonstrates remarkable robustness and stability under increasing view missing rates on the \emph{MIRFLICKR} dataset. Across all three benchmarks, \textsf{\textsf{E2PL}} exhibits a significantly slower performance degradation compared to existing methods, confirming the effectiveness of our efficient prototype tensorization and dynamic contrastive learning strategies. This strong and consistent advantage highlights the generalizability and practical value of \textsf{E2PL} for real-world incomplete multi-view learning scenarios.

\begin{table}[h]
\centering
\setlength{\tabcolsep}{3.5pt}
\caption{Ablation results of \textsf{E2PT} on the \emph{ESPGame} and \emph{IAPRTC12} datasets at $R_V = 30\%$. \textbf{Bold} and \underline{underlined} values indicate the \textbf{best} and \underline{second-best} results, respectively.}
\newcommand{\cmark}{\checkmark}
\begin{tabular}{cccccc}
\toprule
\multicolumn{4}{c}{\textbf{Component}} & \multicolumn{2}{c}{\textbf{mAP}} \\
\cmidrule(lr){1-4} \cmidrule(lr){5-6}
\textbf{+TTPs} & \textbf{+MAPs} & \textbf{+EPT} & \textbf{+DCL} & \emph{ESPGame} & \emph{IAPRTC12} \\
\midrule
      & \cmark  & \cmark & \cmark & 23.87 $\pm$ 0.11 & 24.12 $\pm$ 0.06 \\
\cmark &         &   &  & 26.23 $\pm$ 0.06 &  28.12 $\pm$ 0.03\\
\cmark & \cmark  &      & \cmark & \underline{27.55 $\pm$ 0.03} &  \underline{29.62 $\pm$ 0.05} \\
\cmark & \cmark  & \cmark &      & 27.40 $\pm$ 0.02 & 29.37 $\pm$ 0.04 \\
\cmark & \cmark  & \cmark & \cmark & \textbf{28.29 $\pm$ 0.02}  & \textbf{30.74 $\pm$ 0.04} \\
\bottomrule
\end{tabular}

\label{tab:Ablation2}
\end{table}

\section{Efficiency Analysis}

To rigorously evaluate the effectiveness and efficiency of \textsf{E2PL}, we conduct controlled experiments by replacing the Efficient Prototype Tensorization (EPT) module with MAP~\cite{lee2023multimodal}, MSP~\cite{jang2024towards}, and EPE-P~\cite{chen2024epe} within a unified framework, thereby enabling a direct and systematic comparison of prompt parameterization strategies in terms of predictive performance and parameter complexity.

As reported in Table~\ref{tab:Eff2}, \textsf{E2PL} consistently achieves the highest mAP on both \emph{ESPGame} (28.29) and \emph{IAPRTC12} (30.74), substantially outperforming MAP (27.12/29.78), MSP (26.78/29.19), and EPE-P (27.44/29.90). Importantly, \textsf{E2PL} attains these results with only $\mathcal{O}(n)$ parameter complexity, in stark contrast to the exponential growth of MAP and the cubic complexity of EPE-P. This linear scalability, enabled by the EPT module, allows for expressive and efficient modeling of arbitrary missing-view patterns while preserving computational tractability.

\section{Ablation Studies}

To further validate the necessity and effectiveness of each module in \textsf{E2PL}, we conduct comprehensive ablation studies on the \emph{ESPGame} and \emph{IAPRTC12} datasets, as shown in Table~\ref{tab:Ablation2}. Consistent with the main text, we observe that each component contributes distinct and complementary improvements. Starting from the baseline with only task-tailored prompts (TTPs), the introduction of missing-aware prompts (MAPs) consistently yields a notable boost in mAP, with relative improvements exceeding 10\% on both datasets. Incorporating the efficient prototype tensorization (EPT) module not only further enhances performance, but also dramatically reduces parameter complexity, enabling scalable prompt generation for arbitrary missing-view patterns. The addition of dynamic contrastive learning (DCL) brings further, albeit more moderate, gains, particularly in challenging scenarios with severe view incompleteness. When all modules are integrated, \textsf{E2PL} achieves the best overall results, outperforming all ablated variants by a clear margin. These trends are fully aligned with those observed on \emph{MIRFLICKR} in the main paper, underscoring the robustness and generalizability of our approach. Collectively, these findings confirm that the synergy among TTPs, MAPs, EPT, and DCL is essential for achieving state-of-the-art performance in the IMvMLCIL setting.

\begin{figure}[t]
\centering
  \includegraphics[width=1\columnwidth]{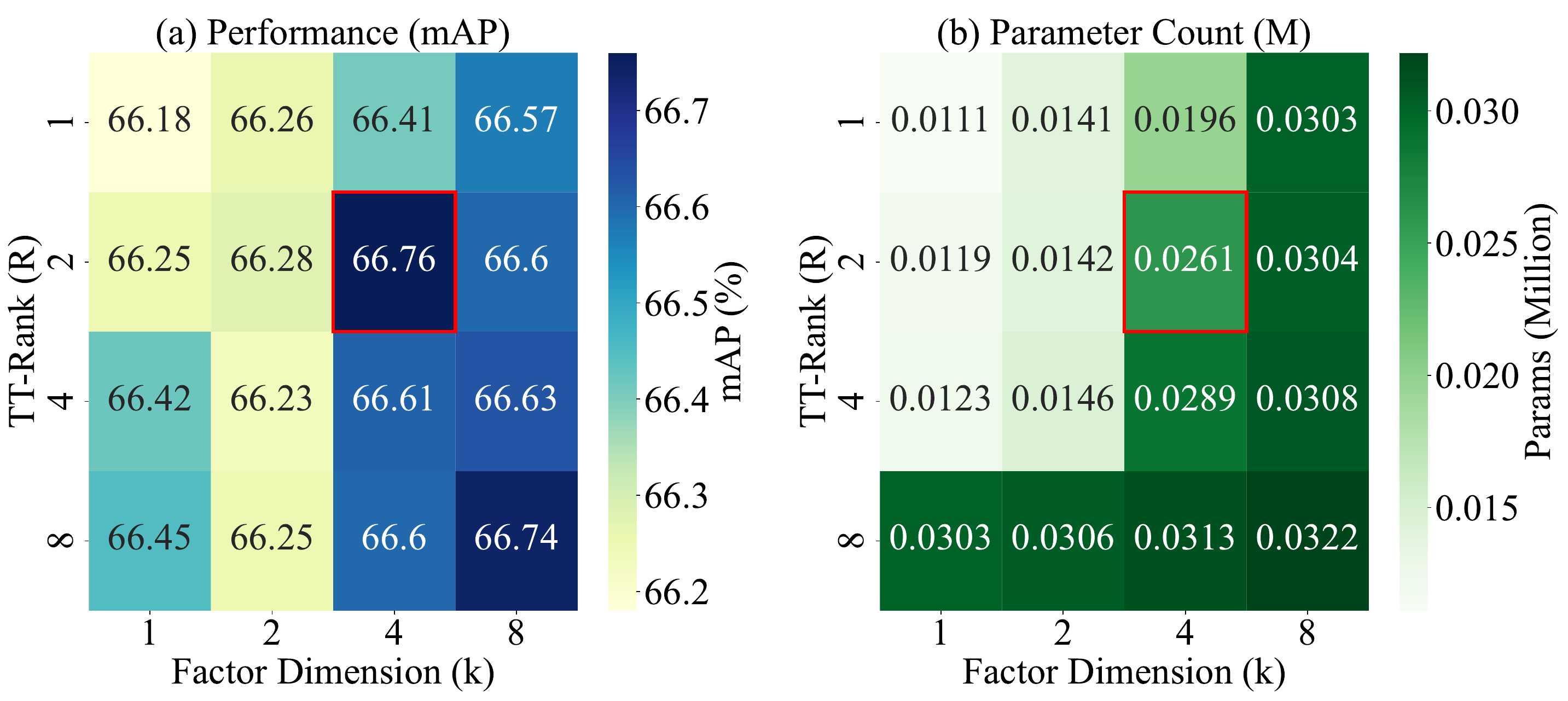}
  \caption{Parameter Sensitivity Analysis of TT-Rank $R$ and Factor Dimension $k$ on the \emph{MIRFLICKR} dataset at $R_V = 30\%$.}
  \label{fig:TT}
\end{figure}

\section{Parameter Evaluation}

We perform a comprehensive sensitivity analysis on the TT-rank $R$ and factor dimension $k$ of the Efficient Prototype Tensorization module, varying both parameters within the set $\{1, 2, 4, 8\}$. As illustrated in Figure~\ref{fig:TT}, increasing $k$ and $R$ generally leads to improved mAP; however, the performance gains become marginal once $k$ and $R$ surpass moderate values. Notably, the configuration $k=4$, $R=2$ achieves the optimal balance between performance and parameter efficiency, yielding the highest mAP ($66.76\%$) with minimal parameter ($0.026 M$) overhead. This configuration strikes a crucial balance: it is expressive enough to capture diverse missing-view patterns, but remains compact to prevent overfitting and excessive resource consumption. Consequently, we adopt $k=4$, $R=2$ in our main experiments, demonstrating that \textsf{E2PL} attains robust and efficient performance with a compact parameterization, which is essential for scalable deployment in practical applications.